\pdfoutput=1
\documentclass[11pt]{article}

\usepackage[preprint]{EMNLP2023}
\usepackage{times}
\usepackage{latexsym}
\usepackage[T1]{fontenc}
\usepackage[utf8]{inputenc}
\usepackage{microtype}
\usepackage{inconsolata}
\usepackage{xspace}
\usepackage{CJKutf8}
\usepackage{booktabs}
\usepackage{multirow}
\usepackage{graphicx}
\usepackage{amsmath}
\usepackage{xcolor}
\usepackage{color, colortbl}

\definecolor{Highlight}{HTML}{C2E8F8}

\newcolumntype{H}{>{\columncolor{Highlight}}c}
\newcommand{\mydataset}{CERD\xspace}
\newcommand{\fulldataset}{\textbf{C}hinese \textbf{E}ssay \textbf{R}hetoric \textbf{D}ataset\xspace}
\newcommand{\rhe}[1]{#1}

\title{\mydataset: A Comprehensive Chinese Rhetoric Dataset for Rhetorical Understanding and Generation in Essays}

\author{
    Nuowei Liu\textsuperscript{1},
    Xinhao Chen\textsuperscript{1},
    Hongyi Wu\textsuperscript{1},
    Changzhi Sun\textsuperscript{1}, \\
    \textbf{Man Lan\textsuperscript{1,2}\thanks{\ Corresponding authors}},
    \textbf{Yuanbin Wu\textsuperscript{1,2}\footnotemark[1],}
    \textbf{Xiaopeng Bai\textsuperscript{2,3},}
    \textbf{Shaoguang Mao\textsuperscript{4},}
    \textbf{Yan Xia\textsuperscript{4}} \\
    \textsuperscript{1}School of Computer Science and Technology, East China Normal University \\
    \textsuperscript{2}Shanghai Institute of AI for Education, East China Normal University \\
    \textsuperscript{3}Department of Chinese Language and Literature, East China Normal University \\
    \textsuperscript{4}Microsoft Research Asia \\
    \texttt{\{nwliu, 51215901006, hongyiwu\}@stu.ecnu.edu.cn}, \texttt{\{czsun.cs\}@gmail.com} \\
    \texttt{\{mlan, ybwu\}@cs.ecnu.edu.cn, xpbai@zhwx.ecnu.edu.cn} \\
    \texttt{\{shaoguang.mao, yanxia\}@microsoft.com}
}

\begin{document}
\maketitle
\begin{abstract}
Existing rhetorical understanding and generation datasets or corpora primarily focus on single coarse-grained categories or fine-grained categories, neglecting the common interrelations between different rhetorical devices by treating them as independent sub-tasks. In this paper, we propose the \fulldataset (\mydataset), consisting of 4 commonly used coarse-grained categories including metaphor, personification, hyperbole and parallelism and 23 fine-grained categories across both form and content levels. \mydataset is a manually annotated and comprehensive Chinese rhetoric dataset with five interrelated sub-tasks. Unlike previous work, our dataset aids in understanding various rhetorical devices, recognizing corresponding rhetorical components, and generating rhetorical sentences under given conditions, thereby improving the author's writing proficiency and language usage skills. Extensive experiments are conducted to demonstrate the interrelations between multiple tasks in \mydataset, as well as to establish a benchmark for future research on rhetoric. The experimental results indicate that Large Language Models achieve the best performance across most tasks, and jointly fine-tuning with multiple tasks further enhances performance. \footnote{Our dataset and code are publicly available at \url{https://github.com/cubenlp/cerd}.}
\end{abstract}

\section{Introduction}
\label{sec:intro}
% Rhetoric, a form of linguistic expression frequently used in Chinese, is often employed in literary works such as novels and prose to enhance the effectiveness and persuasiveness of writing. In the learning process of primary and middle school students, rhetorical devices are a key component of writing skills, with \rhe{metaphor}, \rhe{personification}, \rhe{hyperbole} and \rhe{parallelism} being the most commonly used \citep{chen2019astudy}. Examples of four mentioned coarse-grained categories are shown in Figure \ref{fig:example-rhetoric}. With the advancement of educational technology, many studies explored automatic English or Chinese essay evaluation \citep{rudner2006evaluation,wang2016research,yuan2020automated,zhong2020chinese} where rhetoric is a key component because the use of rhetorical devices in writing reflects the literary quality and language expression ability of an essay \citep{burstein2001system,ishioka2006automated,guo2018attention}.

\begin{figure}[tb]
    \centering
    \includegraphics[width=\hsize]{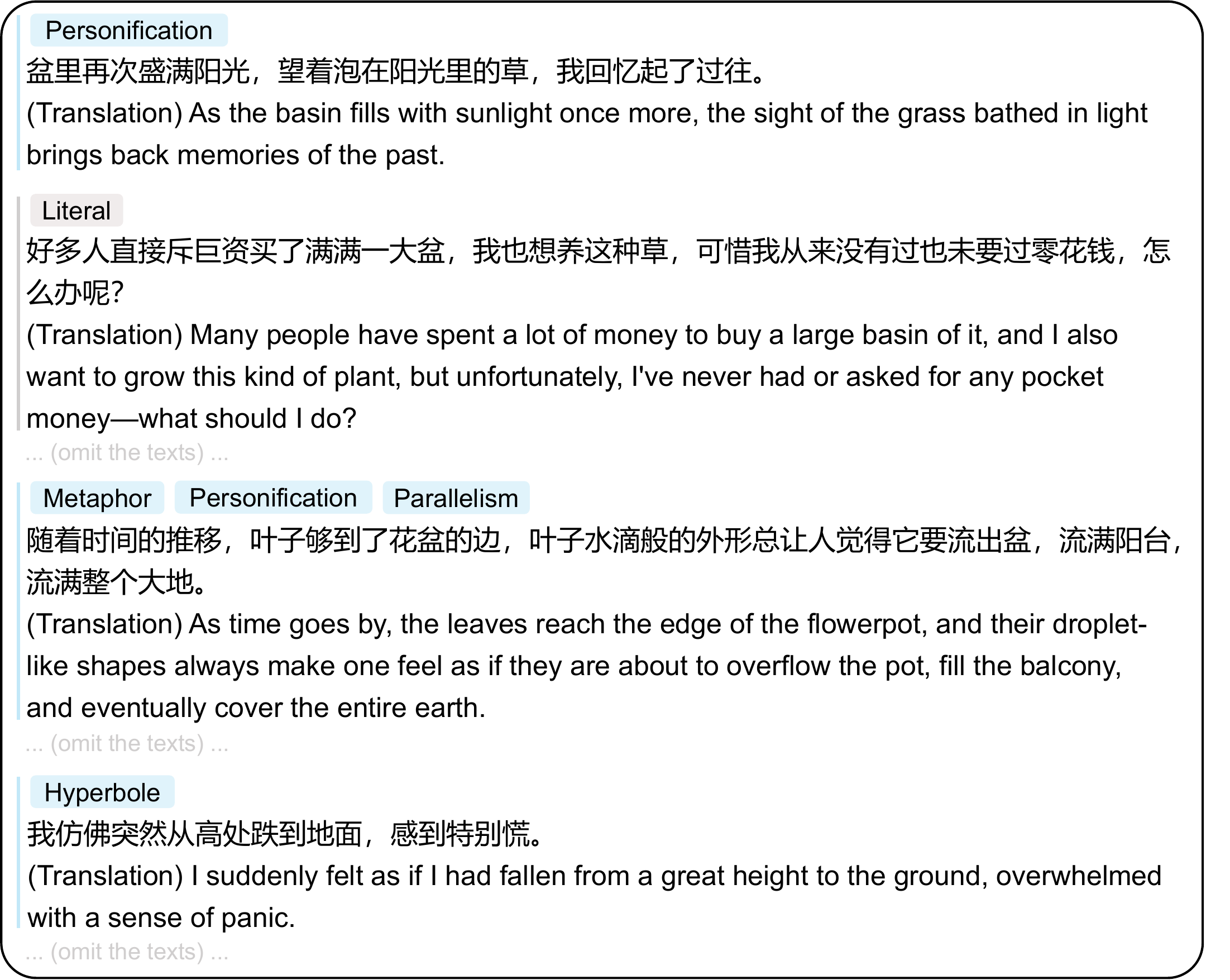}
    \caption{An excerpt from an essay illustrating four commonly used rhetorical devices. It is worth noting that a sentence can employ one or more rhetorical devices, or it can be a literal sentence.}
    \label{fig:example-rhetoric}
\end{figure}

Rhetoric, a form of linguistic expression frequently used in Chinese, is often employed in literary works to enhance the effectiveness and persuasiveness of writing. In the learning process of primary and middle school students, rhetorical devices are a key component of writing skills, with \rhe{metaphor}, \rhe{personification}, \rhe{hyperbole} and \rhe{parallelism} being the most commonly used \citep{chen2019astudy}. Examples of four mentioned coarse-grained categories are shown in Figure \ref{fig:example-rhetoric}. With the advancement of educational technology, several studies explored automatic essay evaluation \citep{wang2016research,yuan2020automated,zhong2020chinese,zhuang2024toree} where rhetoric is a key component because the use of rhetorical devices in writing reflects the literary quality and language expression ability of an essay \citep{burstein2001system,ishioka2006automated}.

% Rhetorical understanding and generation tasks mainly focus on three categories: (1) Category Classification, which classifies the given sentence into coarse-grained or fine-grained categories, (2) Component Extraction, which extracts the rhetorical components in the given sentence as text spans, and (3) Rhetoric Generation, which generates the rhetorical texts based on the given conditions.

% Previous rhetorical understanding related datasets or corpora mainly focused on single coarse-grained categories such as metaphor \citep{shutova2010automatic,li2022nominal} or fine-grained categories such as simile \citep{liu2018neural,chakrabarty2020generating}, a fine-grained category that explicitly uses one specific comparator in a sentence. Furthermore, the common interrelations between different rhetorical devices are not reflected in these datasets because the sub-tasks defined within them are independent. Besides, with the advancements in Large Language Models (LLMs), the task setting of rhetoric generation in the previous work appears to be too simple.
Popular rhetoric benchmarks often excessively focus on a single category of rhetoric and neglect the intrinsic connections between different rhetorical devices, leading to a limited and one-sided understanding of rhetorical phenomena.
For example, \citet{shutova2010automatic} and \citet{li2022nominal} mainly considered metaphors, while \citet{liu2018neural} and \citet{chakrabarty2020generating} only considered similes.
Specifically, \citet{liu2018neural} focused only on similes and the rhetorical components are fixed as tenors and vehicles with a specific comparator in the sentences. Besides, \citet{li2022nominal} introduced a corpus containing metaphorical sentences, treating personification as a type of metaphor. This results in a lack of full utilization of the interrelations between different rhetorical devices.

% To address the challenges, as illustrated in Figure \ref{fig:example-dataset}, we propose the Chinese Essay Rhetoric Dataset (\mydataset), a comprehensive Chinese rhetoric dataset consisting of five sub-tasks: \textbf{(1) Rhetoric Classification (Task RC)}, \textbf{(2) Form Classification (Task FC)}, \textbf{(3) Content Classification (Task CC)}, \textbf{(4) Component Extraction (Task CE)} and \textbf{(5) Rhetoric Generation (Task RG)}. Tasks RC, FC and CC of \mydataset consist of 4 coarse-grained categories and 23 fine-grained categories across two levels. We abstract the types of the rhetorical components of different fine-grained categories, allowing Task CE to be conducted within a unified framework. Furthermore, the annotation was conducted at the essay level, providing Task RG more context for generating rhetorical sentences. Additionally, there are interrelations between multiple tasks in \mydataset, where the annotations from one task can provide additional information to other tasks.

% \begin{figure*}[!htb]
%     \centering
%     \includegraphics[width=\textwidth]{emnlp2023-latex/figs/fig-example-dataset.pdf}
%     \caption{An example of five sub-tasks in \mydataset. An overview of the five tasks is discussed in Section \ref{sec:exp-view}.}
%     \label{fig:example-dataset}
% \end{figure*}

\begin{figure*}[!tb]
    \centering
    \includegraphics[width=0.985\textwidth]{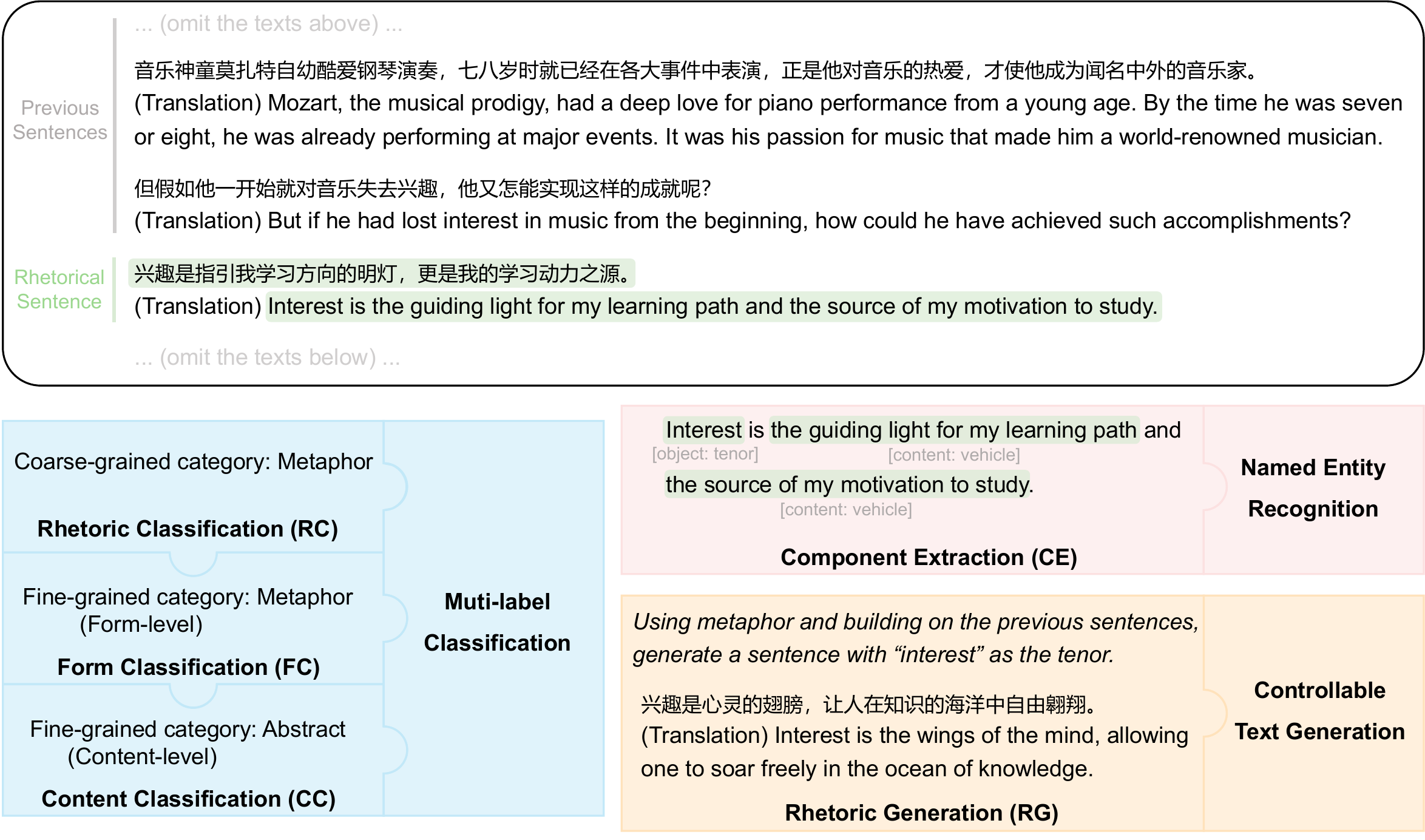}
    \caption{An example of five sub-tasks in \mydataset. An overview of the five tasks is discussed in Section \ref{sec:exp-view}.}
    \label{fig:example-dataset}
\end{figure*}

To address the challenges, as illustrated in Figure \ref{fig:example-dataset}, we propose the \fulldataset (\mydataset), a comprehensive Chinese rhetoric dataset with five sub-tasks, constructed from essays written by primary and middle school students in real educational settings. 
\mydataset addresses the aforementioned limitations in prior work: 
\textbf{Firstly}, our dataset includes 4 coarse-grained categories and 23 fine-grained categories across both form and content levels, providing a broader and deeper perspective for rhetorical understanding.
\textbf{Secondly}, we abstract the types of rhetorical components across different fine-grained categories, enabling their extraction within a unified framework. This approach highlights the intrinsic connections between different rhetorical devices, facilitating a more comprehensive understanding.
\textbf{Thirdly}, unlike previous benchmarks that only required generating parts of the rhetorical components, our dataset provides more context for generating complete rhetorical sentences under certain conditions because the annotation was conducted at the essay level.

The contributions of \mydataset are listed as follows:
\begin{itemize}
    \item We propose the manually annotated \fulldataset (\mydataset) which consists of five interrelated sub-tasks for rhetorical understanding and generation in essays.
    \item Extensive experiments are conducted on \mydataset as a benchmark for future research on rhetoric.
    \item We demonstrate the interrelations between the sub-tasks, highlighting that the annotations from one task can provide additional information to other tasks.
\end{itemize}

\section{Related Work}
\label{sec:related}
Rhetoric studies primarily focus on two categories: understanding and generation.
\paragraph{Rhetoric Datasets}
For rhetorical understanding related datasets, \citet{shutova2010automatic} sampled metaphorical texts from various genres including literature and newspaper articles. \citet{liu2018neural} introduced an annotated Chinese essay corpus focusing on simile. Chinese Literary Grace Corpus (CLGC) presented by \citet{li2022clgc} includes coarse-grained categories of metaphor, personification and parallelism while not further including fine-grained categories or annotations on rhetorical components. For rhetorical generation related datasets, \citet{chakrabarty2020generating} presented a parallel corpus consisting of a large number of similes from collected from Reddit. \citet{li2022nominal} introduced a labeled Chinese Metaphor Corpus (CMC) and a large-scale unlabeled Chinese Literature Corpus (CLC). MAPS-KB \citep{he2023maps} is a million-scale probabilistic simile knowledge base including tenor and vehicle triplets for generating parts of rhetorical components. Distinct from previous work, \mydataset incorporates 4 commonly used coarse-grained categories in a unified framework with 5 interrelated sub-tasks.

\paragraph{Rhetoric Tasks and Approaches}
For rhetorical understanding tasks, \citet{liu2018neural} presented the neural network-based approaches that outperform all rule-based \citep{niculae2013comparison,niculae2013computational,qadir2015learning,qadir2016automatically} and feature-based baselines \citep{li2008computation} on simile related tasks. \citet{zeng2020neural} used the Chinese essay corpus introduced by \citet{liu2018neural} as a benchmark and proposed a cyclic multi-task learning model with a pre-trained BERT \citep{devlin2018bert} encoder that stacks sub-tasks and forms a loop by connecting the last to the first. \citet{wang2022getting} used the same benchmark and present a model that merges the input-side features as a heterogeneous graph and leverages decoding features via distillation. For rhetorical generation tasks, \citet{chakrabarty2020generating} proposed a fine-tuned BART model \citep{lewis2019bart} to generate sentences using similes based on literal sentences. \citet{stowe2021exploring} presented a fine-tuned T5 model \citep{raffel2020exploring} to generate simile sentences in both free-text generation and controllable text generation scenarios. \citet{he2023maps} proposed a framework for large-scale simile knowledge base construction.

\section{Dataset Construction}
\label{sec:dataset}
In this section, we discuss the construction process of \mydataset.The definitions and descriptions of tasks in \mydataset are introduced in Section \ref{sec:exp}.

\subsection{Dataset Overview}
\label{sec:dataset-view}
We collected 503 essays from primary and middle school students' examinations and daily practice, averaging approximately 20.57 sentences and 706.47 tokens per essay. Essays written by students, whose first language is Chinese, are chosen because rhetoric is commonly used in their writing, especially since most of their essays are narrative than argumentative. Furthermore, the essays are written in real educational settings, genuinely reflecting the students' ability to use rhetoric.

\mydataset consists of five tasks, including \textbf{(1) Rhetoric Classification (Task RC)}, \textbf{(2) Form Classification (Task FC)}, \textbf{(3) Content Classification (Task CC)}, \textbf{(4) Component Extraction (Task CE)} and \textbf{(5) Rhetoric Generation (Task RG)}, covering both rhetoric understanding and generation. The annotation was conducted at the essay level, while the results are at the sentence level, except for Task RG.

\subsection{Dataset Annotation}
\label{sec:dataset-anno}

\subsubsection{Dataset Annotation Guidelines}
\label{sec:dataset-guide}
We developed the annotation guidelines based on the linguistic definitions of rhetoric \citep{li2020modern}, categorizing the coarse-grained categories into four types: \rhe{metaphor}, \rhe{personification}, \rhe{hyperbole} and \rhe{parallelism}. We further categorize them into fine-grained categories at both form and content levels. More details are introduced in Appendix \ref{sec:anno-guide-details}.

% TODO Figure

\paragraph{Fine-grained Form-level Categories} The coarse-grained categories are subdivided into 12 fine-grained form-level categories based on the parts of speech or structure of rhetorical components. Fine-grained form-level categories improve the understanding of the structures of rhetorical sentences, facilitating both the analysis of sentence grammar and the extraction of rhetorical components from the sentence.

\paragraph{Fine-grained Content-level Categories} The coarse-grained categories are subdivided into 11 fine-grained content-level categories based on the property of rhetorical components. Fine-grained content-level categories enhance the recognition of the contents and topics of rhetorical sentences, thereby improving the understanding of rhetorical descriptions.

\paragraph{Rhetorical Components} In general, rhetorical components are categorized into three types: connectors, objects and contents. Connectors are used to link the objects and contents or to represent significant markers in a sentence. Objects represent people or things described rhetorically in a sentence. Contents refer to the rhetorical descriptions in a sentence. For different form-level categories, the specific rhetorical components may have various meanings.

\subsubsection{Dataset Annotation Process}
\label{sec:dataset-proc}
During the entire annotation process, as illustrated in Figure \ref{fig:anno-proc} (Appendix \ref{sec:anno-proc-details}), four annotators with backgrounds in Education or Chinese Language and Literature participated. We first developed draft annotation guidelines and conducted a pre-annotation on 50 essays. After assesing the Inter-Annotator Agreements (IAA) \citep{cohen1960coefficient} between the annotators, we refined the draft annotation guidelines. Finally, 503 essays were divided into four batches, with the last 20 essays annotated by Annotator A being the same as the first 20 essays annotated by Annotator B, and so on. These overlapped annotations are used to check the IAA. More details are introduced in Appendix \ref{sec:anno-proc-details}.

\subsection{Dataset Statistics}
\label{sec:dataset-stat}

\subsubsection{Inter-Annotator Agreements}
\label{sec:dataset-iaa}
We use Cohen's Kappa \(\kappa\) \citep{cohen1960coefficient} to evaluate the IAA, defined as Equation \ref{eq:kappa},
\begin{equation}\label{eq:kappa}
    \kappa = \frac{p_{o}-p_{e}}{1-p_{e}}
\end{equation}
where \(p_{o}\) is the empirical probability of agreement on the label assigned to any sample and \(p_{e}\) is the expected agreement when both annotators assign labels randomly. To calculate the IAA for Tasks RC, FC and CC, we use the weighted means of Cohen's Kappa across different categories. For Tasks CE and RG, we remove the tokens that are not part of any rhetorical component and calculate Cohen's Kappa at the token level. The IAA scores across five tasks of \mydataset are shown in Table \ref{tab:iaa}.

\begin{table}[!htb]
    \centering
    \begin{tabular}{@{}ccccc@{}}
      \toprule
      \multicolumn{1}{c}{\multirow{2}{*}{Annotators}} & \multicolumn{4}{c}{Cohen's Kappa \(\kappa\) (\%)} \\ 
      \cmidrule(l){2-5} \multicolumn{1}{c}{} & RC  & FC  & CC  & CE/RG \\ \midrule
        A \& B & 77.67 & 76.01 & 76.87 & 55.89 \\
        B \& C & 59.00 & 58.55 & 58.17 & 45.06 \\
        C \& D & 62.69 & 62.00 & 62.22 & 50.55 \\ \midrule
        Average & 66.45 & 65.54 & 65.76 & 50.50 \\ \bottomrule
      \end{tabular}
    \caption{Inter-Annotator Agreements across five tasks of \mydataset. A, B, C and D denote the four annotators.}
    \label{tab:iaa}
\end{table}

\subsubsection{Dataset Distributions}
\label{sec:dataset-dist}
The distribution of coarse-grained categories across five tasks is shown in Table \ref{tab:coarse-stat}. Sentences using \rhe{metaphor} and \rhe{personification} are more frequent than those employing \rhe{hyperbole} and \rhe{parallelism}, indicating that these are the most commonly rhetorical devices used in students' essays.

\begin{table}[!htb]
    \centering
    \begin{tabular}{cccccc}
    \toprule
      Task & \#Met & \#Per & \#Hyp & \#Par & \#Lit \\ \midrule
        RC & 509 & 220 & 130 & 150 & 150 \\
        FC & 524 & 229 & 132 & 151 & 150 \\ 
        CC & 522 & 221 & 130 & 151 & 150 \\ 
        CE & 572 & 271 & 136 & 152 & 150 \\ 
        RG & 449 & 260 & 135 & 0 & 0 \\ \bottomrule
    \end{tabular}
    \caption{Distribution of coarse-grained categories across five tasks. "Met", "Per", "Hyp", "Par", "Lit" refer to metaphor, personification, hyperbole, parallelism and literal, respectively. A sentence can employ several rhetorical devices, which are not counted redundantly in the Task RC. Furthermore, Task RG excludes all sentences that use parallelism and literal sentences.}
    \label{tab:coarse-stat}
\end{table}

\begin{figure*}[!htb]
    \centering
    \includegraphics[width=\textwidth]{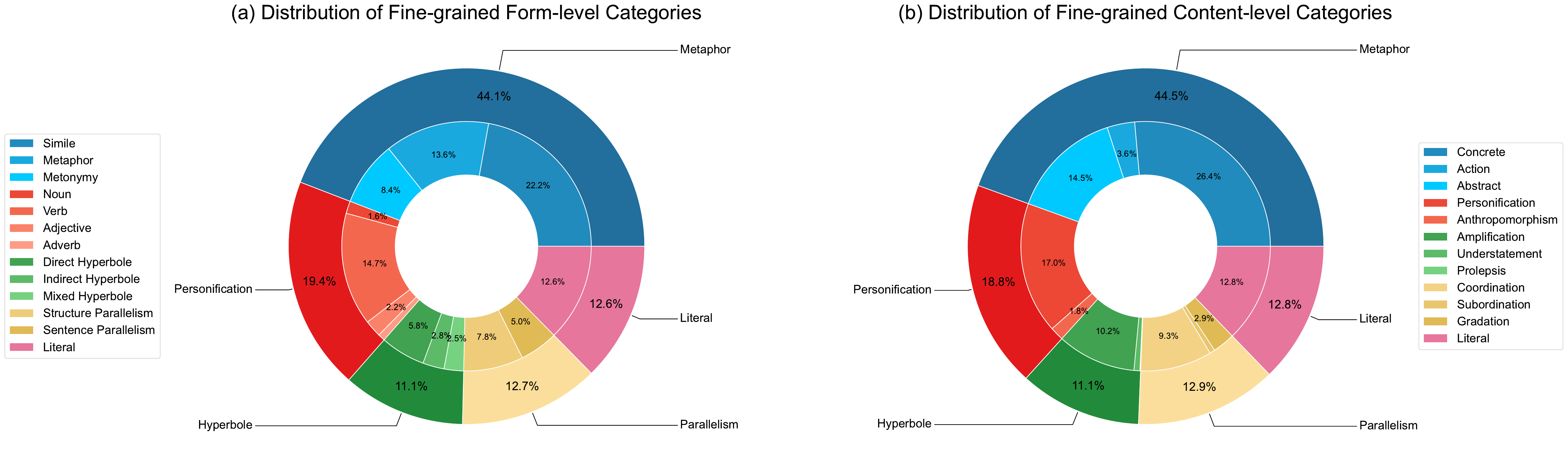}
    \caption{Distribution of fine-grained categories is illustrated in Figure (a) for form-level categories and in Figure (b) for content-level categories.}
    \label{fig:dataset-dist}
\end{figure*}

Figure \ref{fig:dataset-dist} (a) shows the distribution of fine-grained form-level categories, with \rhe{simile} and \rhe{verb} being the most common. We also assess the distribution of fine-grained content-level categories, displayed in Figure \ref{fig:dataset-dist} (b), illustrating that the content categories of concrete and personification are the most frequently used.

\section{Experiments}
\label{sec:exp}
\subsection{Tasks Overview}
\label{sec:exp-view}
\mydataset includes five tasks, covering multiple task types such as multi-label classification, named entity recognition and controllable text generation, providing comprehensive support for rhetorical understanding and generation.

\paragraph{Rhetoric/Form/Content Classification} Tasks RC/FC/CC are multi-label classification problems. Given a sentence \(x\) as input, a model is asked to predict which rhetorical devices \(y \subset  Y\) the sentence employs, where the set \(Y\) denotes all the possible categories in a task. In particular, a sentence may employ multiple rhetorical devices. Therefore, \(|y|\) should satisfy \(1 \leq |y| \leq |Y|\). For Task RC, there are 5 possible coarse-grained categories, including the case of literal sentences. For Task FC, there are 13 possible fine-grained form-level categories, including the case of literal sentences. For Task CC, there are 12 possible fine-grained content-level categories, including the case of literal sentences. 

\paragraph{Component Extraction} Task CE is a named entity recognition problem. Given a sentence \(x\) with \(N\) tokens as input, a model is expected to extract all the possible rhetorical components \(y\) in the sentence, where \(y = \{S_{\text{literals}}, S_{\text{connectors}}, S_{\text{objects}}, S_{\text{contents}}\}\) is a tuple. The set \(S\) consists of multiple ordered pairs \((i, j)\), where \(1\leq i \leq j \leq N\) denotes the indices of the literal or rhetorical components in the sentence.

\paragraph{Rhetoric Generation} Task RG is a controllable text generation problem. For an essay with \(N\) sentences, given the preceding context with at most \(k\) consecutive sentences \(s = \{s_{i-k}, \ldots, s_{i-2}, s_{i-1}\}\), the objects of the \(i\)-th sentence, and the coarse-grained categories the \(i\)-th sentence employs as inputs, a model is asked to generate the sentence \(s_{i}\) satisfying the conditions, where \(1\leq i \leq N, k=\min\{k, i-1\}\).

\paragraph{\textit{Interrelations between the Tasks}} There are interrelations between multiple tasks in \mydataset, where the annotations from one task can provide additional information to other tasks. Tasks FC and CC rely on the coarse-grained categories provided by Task RC. Furthermore, Task CE relies on the fine-grained form-level categories from Task FC. Additionally, Task RG relies on the coarse-grained categories from Task RC and the rhetorical components extracted by Task CE.

\subsection{Baselines and Evaluation Metrics}
\label{sec:exp-base}

\paragraph{Baselines} We evaluate RoBERTa \citep{liu2019roberta}, a BERT-based \citep{devlin2018bert} pre-trained model on Task RC, FC, CC and CE. Furthermore, we test LLMs such as GPT-3.5 \citep{openai2022chatgpt}, GPT-4 \citep{achiam2023gpt} and Qwen1.5 \cite{bai2023qwen} on all the tasks. In particular, for RoBERTa, we choose \(\text{RoBERTa}_{\text{BASE}}\)\footnote{\url{https://huggingface.co/uer/chinese_roberta_L-12_H-768}} pre-trained on Chinese corpus CLUECorpusSamll \citep{xu2020cluecorpus2020}. For GPT-3.5 and GPT-4, we use gpt-3.5-turbo-0125 and gpt-4-turbo-2024-04-09 respectively. For Qwen1.5, we adopt both zero-shot learning and LoRA \citep{hu2021lora} fine-tuning for all the tasks. Details of the experimental setups are provided in Appendix \ref{sec:exp-setups}.

\paragraph{Evaluation Metrics} To evaluate Tasks RC, FC, CC and CE, we utilize the metrics such as Exact Match, Precision, Recall and F1 score. In particular, seqeval \citep{ramshaw1999text,seqeval2018hiroki}, a framework for sequence labeling evaluation, is used to assess Task CE. To evaluate Task RG, we adopt automatic evaluation metrics such as BLEU \citep{papineni2002bleu}, ROUGE \citep{lin2004rouge}, PPL \citep{jelinek1977perplexity} and BERTScore \citep{zhang2019bertscore}. We also use LLMs like GPT-4o \citep{openai2024gpt4o} to evaluate the quality of the models' generations. Specifically, we design two LLM-based evaluation metrics: Single-answer Rating and Pairwise Ranking. The Single-answer Rating metric asks the LLM to rate the generations on a scale from 1 to 5. The Pairwise Ranking metric asks the LLM to compare the generated sentences with the original ones written in the essays. To systematically assess the quality of the generated sentences, 10 individuals are invited to evaluate the results of Task RG, with 2 individuals working together as a group for each model.

\subsection{Results and Analysis}
\label{sec:exp-res}

\subsubsection{Rhetoric Classification}
\label{sec:exp-rc}
As shown in Table \ref{tab:exp-rc}, Qwen1.5-7B with multi-task fine-tuning outperforms all other models in classifying coarse-grained categories. Besides, RoBERTa fine-tuned on the task surpasses all the LLMs in zero-shot performance but scores slightly lower than Qwen1.5-7B with single-task fine-tuning.

\begin{table*}[!htb]
    \centering
    \small
    \begin{tabular}{lcccHccH}
     \toprule
     Models & EM & micro-P & micro-R & micro-F1 & macro-P & macro-R & macro-F1  \\ \midrule
      RoBERTa & 63.31 & 72.40 & \underline{76.81} & 74.54 & 68.75 & \underline{69.00} & 68.36 \\ \midrule
      GPT-3.5 & 20.16 & 37.39 & 64.26 & 47.27 & 30.61 & 51.95 & 36.10 \\
      GPT-4 & 54.44 & 61.46 & 70.34 & 65.50 & 54.21 & 63.36 & 57.11 \\
      Qwen1.5-7B & 27.82 & 40.54 & 68.44 & 50.92 & 31.43 & 54.35 & 38.69 \\
      \quad w/ single-task FT & \underline{71.77} & \underline{77.25} & 74.90 & \underline{76.06} & \underline{73.05} & 68.29 & \underline{70.27} \\
      \quad w/ multi-task FT & \textbf{75.40} & \textbf{80.56} & \textbf{77.19} & \textbf{78.83} & \textbf{76.71} & \textbf{70.02} & \textbf{72.68} \\ \bottomrule
    \end{tabular}
    \caption{Results (in \%) of Rhetoric Classification Task.}
    \label{tab:exp-rc}
\end{table*}

The experimental results indicate that the BERT-based model outperforms LLMs, as the gap between coarse-grained categories is significantly larger than the gap between fine-grained categories.

\subsubsection{Form Classification}
\label{sec:exp-fc}
As shown in Table \ref{tab:exp-fc}, for a more complicated multi-label classification problem, RoBERTa performs competitively with LLMs. In particular, RoBERTa outperforms Qwen1.5-7B with both single-task fine-tuning and multi-task fine-tuning on the micro-F1 score. However, Qwen1.5-7B with fine-tuning performs significantly better than RoBERTa on the macro-F1 score, while Qwen1.5-7B with zero-shot approaches the performance of RoBERTa and GPT-4 in zero-shot settings.

\begin{table*}[!htb]
    \centering
    \small
    \begin{tabular}{lcccHccH}
     \toprule
     Models & EM & micro-P & micro-R & micro-F1 & macro-P & macro-R & macro-F1  \\ \midrule
      RoBERTa & \underline{50.81} & \textbf{76.63} & \underline{52.03} & \textbf{61.98} & \textbf{86.13} & 29.93 & 33.92 \\ \midrule
      GPT-3.5 & 2.42 & 12.86 & 29.89 & 17.98 & 33.85 & 25.97 & 20.02 \\
      GPT-4 & 24.60 & 33.06 & 43.91 & 37.72 & 37.48 & \underline{30.78} & 30.39 \\
      Qwen1.5-7B & 5.24 & 14.39 & 35.42 & 20.47 & 20.13 & 25.22 & 28.99 \\
      \quad w/ single-task FT & 41.94 & 47.98 & 43.91 & 45.86 & \underline{52.09} & 24.92 & \underline{40.20} \\
      \quad w/ multi-task FT & \textbf{54.03} & \underline{59.60} & \textbf{54.98} & \underline{57.20} & 51.46 & \textbf{31.81} & \textbf{55.04} \\ \bottomrule
    \end{tabular}
    \caption{Results (in \%) of Form Classification Task.}
    \label{tab:exp-fc}
\end{table*}

\subsubsection{Content Classification}
\label{sec:exp-cc}
As shown in Table \ref{tab:exp-cc}, RoBERTa outperforms all the LLMs on all metrics except for macro-Recall and macro-F1, while Qwen1.5-7B with multi-task fine-tuning approaches the performance of RoBERTa. Notably, GPT-4 surpasses all other baselines on the macro-F1 score by approximately 15\% compared to the second best model.

\begin{table*}[!htb]
    \centering
    \small
    \begin{tabular}{lcccHccH}
     \toprule
     Models & EM & micro-P & micro-R & micro-F1 & macro-P & macro-R & macro-F1  \\ \midrule
      RoBERTa & \textbf{54.44} & \textbf{67.95} & \textbf{59.77} & \textbf{63.60} & \textbf{75.55} & 40.44 & 43.49 \\ \midrule
      GPT-3.5 & 2.82 & 16.35 & 32.71 & 21.80 & 21.34 & 31.76 & 31.80 \\
      GPT-4 & 12.50 & 23.84 & 28.95 & 26.15 & 25.79 & 29.31 & \textbf{58.26} \\
      Qwen1.5-7B & 2.42 & 16.90 & 35.71 & 22.95 & 18.69 & 35.95 & 33.89 \\
      \quad w/ single-task FT & 46.77 & 51.21 & 47.74 & 49.42 & \underline{66.49} & 35.92 & 36.96 \\
      \quad w/ multi-task FT & \underline{53.63} & \underline{59.68} & \underline{56.77} & \underline{58.19} & 55.19 & \textbf{42.27} & \underline{43.85} \\ \bottomrule
    \end{tabular}
    \caption{Results (in \%) of Content Classification Task.}
    \label{tab:exp-cc}
\end{table*}

The experimental results of Tasks FC and CC on the macro-F1 scores highlight that LLMs are more capable of understanding imbalanced fine-grained categories than BERT-based model. This is possibly because LLMs learn the concepts and differences of various categories through prompts, which will be further discussed in Appendix \ref{sec:prompt}.

\begin{figure}[!htb]
    \centering
    \includegraphics[width=\hsize]{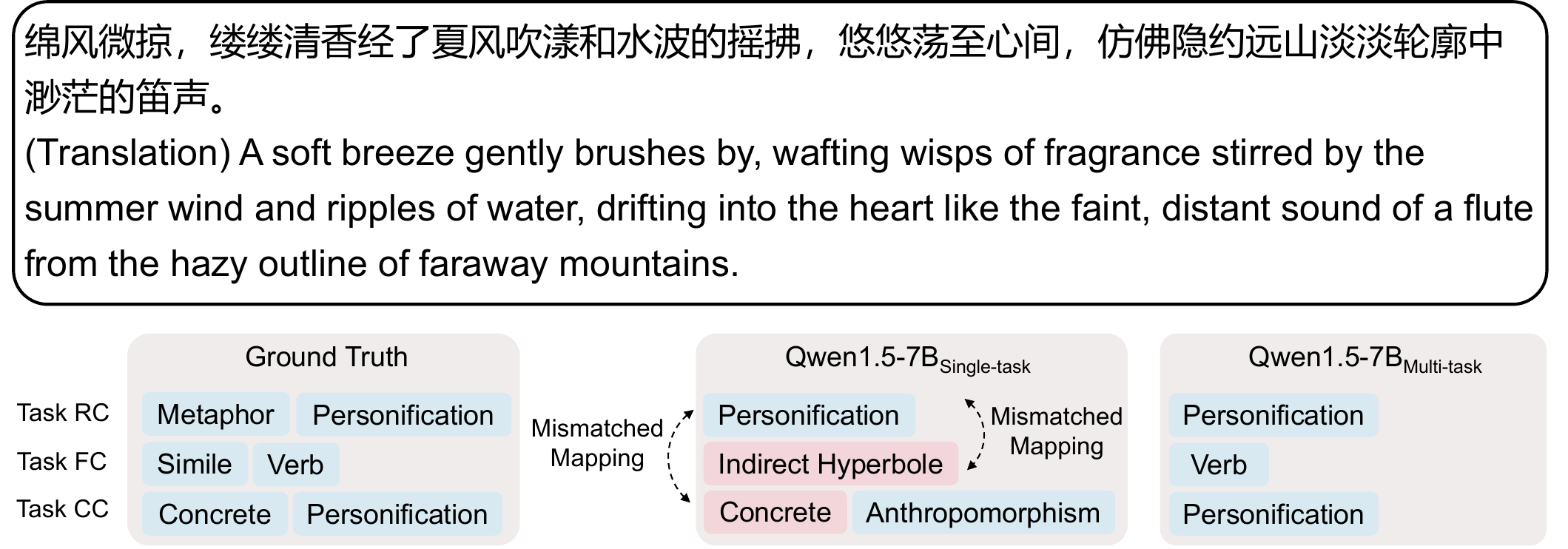}
    \caption{Case study on Rhetoric Classification Task, Form Classification Task and Content Classification Task. A mismatched mapping refers to a fine-grained category that does not belong to its predicted corresponding coarse-grained category.}
    \label{fig:case-fc-cc}
\end{figure}

Furthermore, compared to Task RC, Qwen1.5-7B with multi-task fine-tuning surpasses the model fine-tuned on the single task, demonstrating that it learns the interrelations between different tasks. A possible explanation is that the model learns the mappings of coarse-grained and fine-grained categories through multi-task fine-tuning. As illustrated in Figure \ref{fig:case-fc-cc}, the given sentence employs both metaphor and personification, while Qwen1.5-7B with single-task fine-tuning classifies it as personification. Additionally, for Task FC, the model predicts the sentence as indirect hyperbole, which is a fine-grained category of hyperbole rather than personification. The mismatched mapping between coarse-grained and fine-grained categories also occurs in Task CC, indicating that the model fails to establish the correct mappings through single-task fine-tuning. Further analysis of the mappings between categories is discussed in Section \ref{sec:exp-ablation-rc}.

\subsubsection{Component Extraction}
\label{sec:exp-ce}
As shown in Table \ref{tab:exp-ce}, Qwen1.5-7B with multi-task fine-tuning is competitive with RoBERTa on both the micro-F1 and macro-F1 scores. Additionally, GPT-4 with zero-shot achieves the best performance on Recall metrics.

\begin{figure}[!htb]
    \centering
    \includegraphics[width=\hsize]{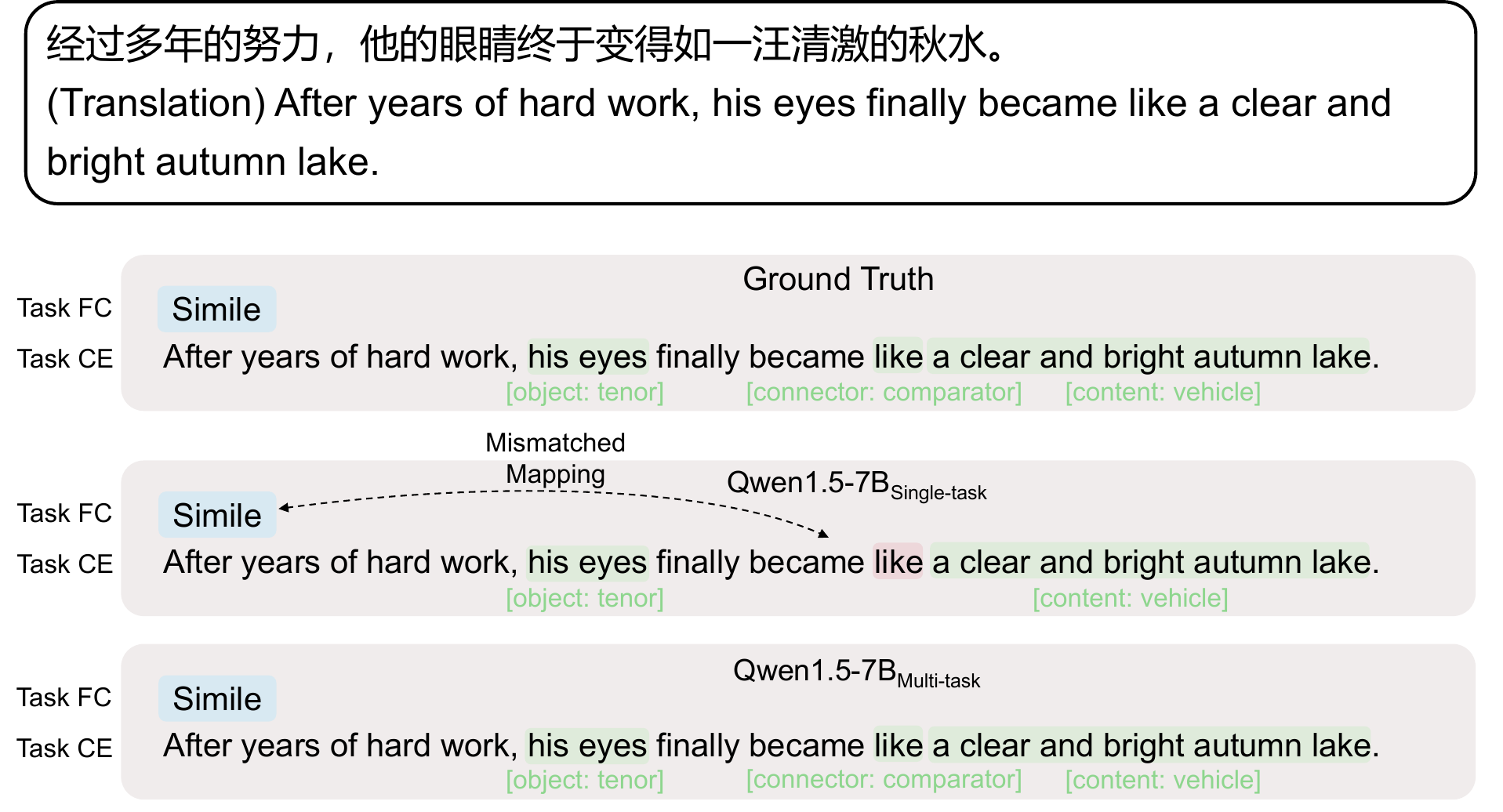}
    \caption{Case study on Component Extraction Task. A mismatched mapping refers to the extracted rhetorical components that do not fully satisfy the requirements of the predicted corresponding fine-grained form-level category.}
    \label{fig:case-ce}
\end{figure}

As illustrated in Figure \ref{fig:case-ce}, the fine-grained form-level category of the given sentence is simile, which requires comparator, tenor and vehicle as its rhetorical components. Qwen1.5-7B with single-task fine-tuning fails to extract the comparator from the sentence, even though the model classifies it as a simile sentence. Further analysis of mappings between rhetorical components and fine-grained form-level categories is discussed in Section \ref{sec:exp-ablation-fc}.

\begin{table*}[!htb]
    \centering
    \small
    \begin{tabular}{lcccHccH}
     \toprule
     Models & Acc & micro-P & micro-R & micro-F1 & macro-P & macro-R & macro-F1  \\ \midrule
      RoBERTa & \textbf{89.23} & 38.84 & \underline{40.61} & \textbf{39.70} & 42.26 & \underline{43.49} & \underline{42.83} \\ \midrule
      GPT-3.5 & 52.09 & 10.01 & 29.98 & 15.01 & 12.66 & 29.88 & 17.07 \\
      GPT-4 & 71.20 & 29.10 & \textbf{44.40} & 35.16 & 30.01 & \textbf{46.73} & 36.51 \\
      Qwen1.5-7B & 56.17 & 11.34 & 33.40 & 16.93 & 11.41 & 36.39 & 17.20 \\
      \quad w/ single-task FT & \underline{83.82} & \underline{40.82} & 32.07 & 35.92 & \textbf{51.72} & 31.63 & 37.14 \\
      \quad w/ multi-task FT & 82.64 & \textbf{41.81} & 37.76 & \underline{39.68} & \underline{46.21} & 40.32 & \textbf{43.00} \\ \bottomrule
    \end{tabular}
    \caption{Results (in \%) of Component Extraction Task.}
    \label{tab:exp-ce}
\end{table*}

\subsubsection{Rhetoric Generation}
\label{sec:exp-rg}
As shown in Table \ref{tab:exp-rg-auto}, Qwen1.5-7B and GPT-4 with zero-shot exhibit competitive performances across multiple metrics. Specifically, for automatic evaluation metrics, Qwen1.5-7B achieves the best performance on BLEU-2 and PPL, while GPT-4 surpasses other baselines on BLEU-4, ROUGE-L and BERTScore.

As shown in Table \ref{tab:exp-rg-llm}, for both LLM-based evaluation metrics and human evaluations, GPT-4 achieves the highest Single-answer rating score, indicating its capability to generate fluent and expressive rhetorical sentences. Furthermore, Qwen1.5-7B performs the best on the Pairwise Ranking metric, demonstrating that 69.23\% of its generated rhetorical sentences are better than the references in essays. However, it is worth noting that compared to Qwen1.5-7B with zero-shot, the model fine-tuned on Task RG or multi-task performs worse. A potential reason is that the model overfits on the training set and therefore loses its generalization capability. 

\begin{table*}[!htb]
    \centering
    \small
    \begin{tabular}{lccccccc}
     \toprule
        & BLEU-2 & BLEU-4 & ROUGE-L & PPL & \(P_{\text{BERT}}\) & \(R_{\text{BERT}}\) & \(F_{\text{BERT}}\) \\ 
    Models & (\%) \(\uparrow\) & (\%) \(\uparrow\) & (\%) \(\uparrow\) & \(\downarrow\) & (\%) \(\uparrow\) & (\%) \(\uparrow\) & (\%) \(\uparrow\) \\ \midrule
     GPT-3.5 & 6.55 & 3.23 & \underline{19.13} & 81.10 & \underline{65.42} & 61.52 & \underline{63.30}  \\
      GPT-4 & 6.82 & \textbf{3.43} & \textbf{20.33} & \underline{45.79} & \textbf{66.32} & \textbf{62.62} & \textbf{64.30} \\
      Qwen1.5-7B & \textbf{8.27} & \underline{3.24} & 17.43 & \textbf{45.17} & 63.63 & \underline{62.44} & 62.93  \\
      \quad w/ single-task FT & \underline{6.96} & 2.77 & 14.74 & 154.39 & 61.51 & 58.83 & 60.01 \\
      \quad w/ multi-task FT & 5.83 & 1.69 & 14.61 & 125.96 & 59.94 & 58.72 & 59.23 \\ \bottomrule
    \end{tabular}%
    \caption{Results of Rhetoric Generation Task using automatic evaluation metrics.}
    \label{tab:exp-rg-auto}
\end{table*}

\begin{table*}[!htb]
    \centering
    \small
    \begin{tabular}{lcccc}
     \toprule
     \multicolumn{1}{c}{} & \multicolumn{2}{c}{LLM-based Evaluation} & \multicolumn{2}{c}{Human Evaluation} \\
     \cmidrule(lr){2-3} \cmidrule(lr){4-5}
    Models & Rating & Ranking (\%) & Rating & Ranking (\%) \\ \midrule
     GPT-3.5 & 4.01 & 59.17 & 3.06 & 33.72 \\
      GPT-4 & \textbf{4.61} & \underline{66.27} & \textbf{3.85} & 37.21 \\
      Qwen1.5-7B & \underline{4.14} & \textbf{69.23} & \underline{3.79} & \textbf{40.70}   \\
      \quad w/ single-task FT & 1.67 & 19.53 & 3.46 & 37.21  \\
      \quad w/ multi-task FT & 1.97 & 29.59 & 3.56 & \underline{39.53} \\ \bottomrule
    \end{tabular}%
    \caption{Results of Rhetoric Generation Task using LLM-based evaluation metrics and human evaluations. "Rating" refers to Pairwise-answer Rating, a score from 1 to 5. "Ranking" refers to Pairwise Ranking, indicating the percentage of generated sentences better than the references.}
    \label{tab:exp-rg-llm}
\end{table*}

\begin{figure}[!tb]
    \centering
    \includegraphics[width=\hsize]{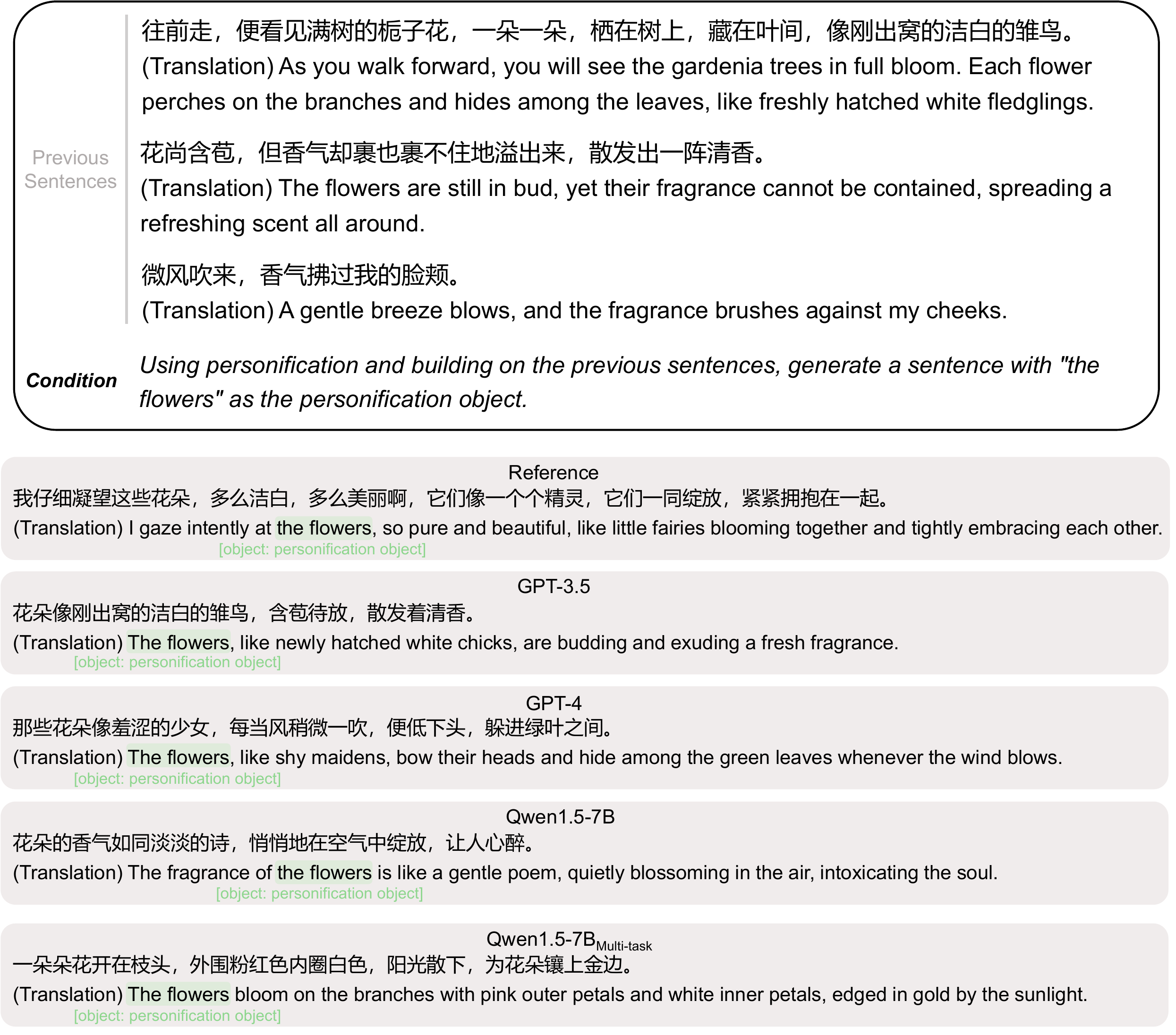}
    \caption{Case study on Rhetoric Generation Task. The prompt is originally in Chinese, with the English translation provided for illustration.}
    \label{fig:case-rg}
\end{figure}

An example of rhetorical sentences generated by various models is illustrated in Figure \ref{fig:case-rg}, indicating that GPT-3.5, GPT-4 and Qwen1.5-7B generate the rhetorical sentences satisfying the given conditions. Besides, the generation closely relates to the preceding context. For example, GPT-3.5 and Qwen1.5-7B mention the fragrance of flowers that appeared earlier in the text, while GPT-4 references the previously mentioned breeze.

\section{Discussion}
\label{sec:discuss}
\subsection{Effect of Rhetoric Classification Task}
\label{sec:exp-ablation-rc}
As mentioned in Section \ref{sec:exp-view} and Section \ref{sec:exp-cc}, Task RC provides information on coarse-grained categories, while Tasks FC and CC require the model to classify sentences at fine-grained levels. Intuitively, it is much more complicated for a model to directly solve Tasks FC and CC because the number of fine-grained categories is larger than that of coarse-grained ones. Therefore, learning the mappings between coarse-grained categories and their corresponding fine-grained categories may help the model solve Tasks FC and CC.

\begin{figure}[!htb]
    \centering
    \includegraphics[width=\hsize]{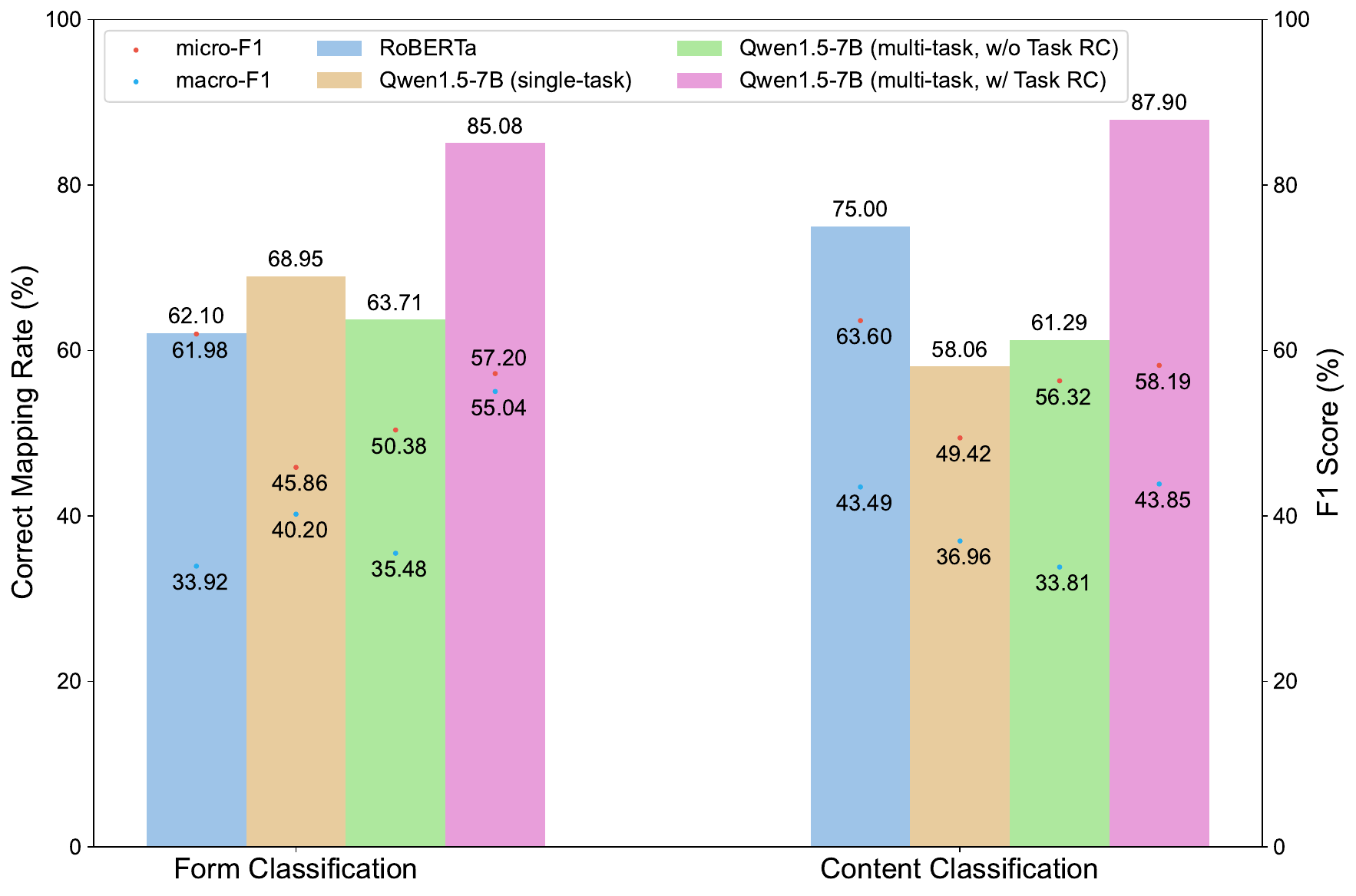}
    \caption{Effect of Task RC during multi-task fine-tuning. The bars represent the correct mapping rates, while the points represent the F1 scores.}
    \label{fig:ablation-rc}
\end{figure}

We define the correct mapping rate as the percentage of instances where a model correctly maps all coarse-grained categories in Task RC to their corresponding fine-grained form-level or content-level categories in Tasks FC or CC. As displayed in Figure \ref{fig:ablation-rc}, RoBERTa and Qwen1.5-7B fine-tuned on the single task show similar but relatively low performance on correct mapping rates. When Task RC is removed from the multi-task fine-tuning stage, there are no significant differences on correct mapping rates compared to Qwen1.5-7B with single-task fine-tuning. However, reintroducing Task RC data during multi-task fine-tuning significantly improves the performance of Qwen1.5-7B on correct mapping rate. Therefore, the experiment demonstrates the effect of Task RC on the mappings between coarse-grained and fine-grained categories.

\subsection{Effect of Form Classification Task}
\label{sec:exp-ablation-fc}
\begin{figure}[!htb]
    \centering
    \includegraphics[width=0.75\hsize]{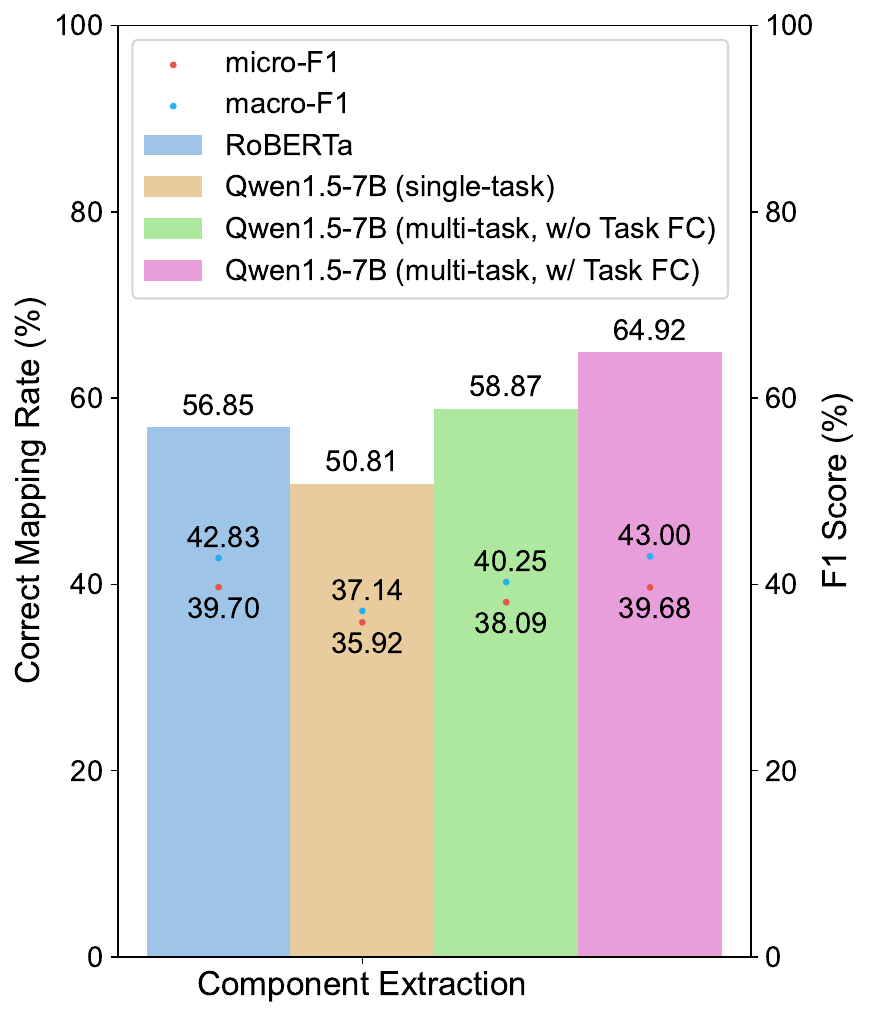}
    \caption{Effect of Task FC during multi-task fine-tuning. The bars represent the correct mapping rates, while the points represent the F1 scores.}
    \label{fig:ablation-fc}
\end{figure}

Similar to the correct mapping rate in Section \ref{sec:exp-ablation-rc}, the correct mapping rate of Task CE is defined as the percentage of instances where a model extracts all the necessary rhetorical components in a given sentence according to its form-level categories. As shown in Figure \ref{fig:ablation-fc}, compared to RoBERTa and Qwen1.5-7B fine-tuned without Task FC, Qwen1.5-7B with multi-task fine-tuning improves the correct mapping rate. The results demonstrate the importance of Task FC in extracting correct rhetorical components from the sentences.

\subsection{Error Analysis}
\label{sec:exp-error}
\begin{figure}[!htb]
    \centering
    \includegraphics[width=\linewidth]{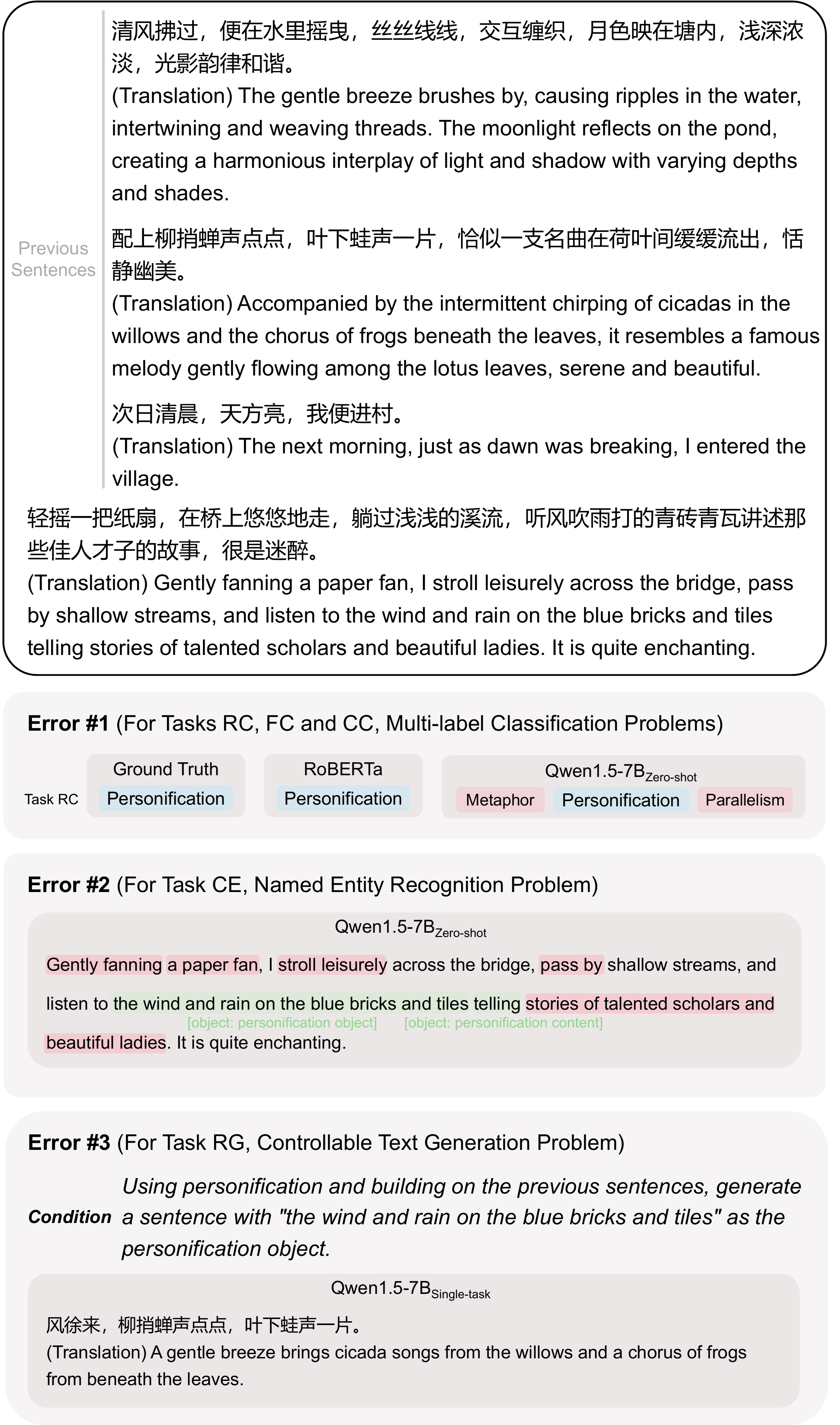}
        \caption{Analysis of common and consistent error patterns made by models.}
    \label{fig:error}
\end{figure}

An error study on common and consistent error patterns made by models, especially Large Language Models, is illustrated in Figure \ref{fig:error}. 

For Tasks RC, FC, and CC, aside from the mismatched mapping discussed in Section \ref{sec:exp-cc} and Section \ref{sec:exp-ablation-rc}, a common error pattern is that models with zero-shot setting tend to classify a given sentence into an excessively large number of categories. For instance, as shown in Figure \ref{fig:error}, the given sentence employs personification, while Qwen1.5-7B with zero-shot predicts it using three rhetorical devices. 

For Task CE, similar to multi-label classification problems, an error is mismatched mapping, as discussed in Section \ref{sec:exp-ce} and Section \ref{sec:exp-ablation-fc}. Additionally, models with zero-shot tend to extract longer spans of rhetorical components or incorrectly identify literal parts as rhetorical components.

For Task RG, one error pattern is that the generated sentences may fail to use the required rhetorical devices with the specified objects. Another common error is that the generated sentences may have little or no relation to the preceding context. Although the cicadas and frogs in the generated sentence in Figure \ref{fig:error} are relevant to the preceding context, Qwen1.5-7B still fails to generate the specified object correctly.

\section{Conclusion}
\label{sec:conclusion}
In this paper, we propose the \fulldataset (\mydataset), a comprehensive Chinese rhetoric dataset consisting of five sub-tasks. We conduct extensive experiments as a benchmark for future research on rhetoric. The experimental results indicate that both GPT-4 and Qwen1.5-7B with fine-tuning are superior baseline models, achieving competitive performances across multiple sub-tasks. Furthermore, we demonstrate the interrelations between different sub-tasks in \mydataset and the significance of task settings.

\section*{Limitations}
\label{sec:limitations}
The data collected to construct \mydataset comes from real educational settings. Although it does not affect the recognition and understanding of rhetoric, there may inevitably be some typographical errors due to the limited language proficiency of primary and middle school students.

\section*{Ethics Statement}
\label{sec:ethics-state}
All the participating annotators were compensated for their contributions, with each annotator's hourly wage being approximately 45\% higher than the local minimum wage. Additionally, all the essays in \mydataset have been authorized for use. Moreover, to protect the privacy of the authors, we adopted data anonymization in \mydataset, removing all personal information related to them.

\section*{Acknowledgements}
\label{sec:ack}
We appreciate the support from National Natural Science Foundation of China with the Main Research Project on Machine Behavior and Human Machine Collaborated Decision Making Methodology(72192820 \& 72192824), Fundamental Research Funds for the Central Universities (2024QKT004), Pudong New Area Science \& Technology Development Fund (PKX2021-R05), Science and Technology Commission of Shanghai Municipality (22DZ2229004), and Shanghai Trusted Industry Internet Software Collaborative Innovation Center.

\bibliography{main}
% \bibliographystyle{acl_natbib}

% \clearpage
\appendix
\section{Dataset Annotation Details}
\label{sec:anno-details}

\subsection{Details of Dataset Annotation Guidelines}
\label{sec:anno-guide-details}
The annotation guidelines for form-level and content-level categories in \mydataset is shown in Table \ref{tab:guideline-fc} and \ref{tab:guideline-cc} respectively. We subdivide the coarse-grained categories into fine-grained form-level and content-level categories based on specific criteria. Specifically, the fine-grained form-level categories include:
\begin{itemize}
    \item For metaphor, it is subdivided into simile, metaphor and metonymy.
    \item For personification, it is subdivided into noun, verb, adjective and adverb.
    \item For hyperbole, it is subdivided into direct hyperbole, indirect hyperbole and mixed hyperbole.
    \item For parallelism, it is subdivided into structure parallelism and sentence parallelism.
    \end{itemize}

Besides, the fine-grained content-level categories include:
\begin{itemize}
    \item For metaphor, it is subdivided into concrete, action and abstract.
    \item For personification, it is subdivided into personification and anthropomorphism.
    \item For hyperbole, it is subdivided into amplification, understatement and prolepsis.
    \item For parallelism, it is subdivided into coordination, subordination and gradation.
\end{itemize}

\begin{table*}[!htb]
    \centering
    \resizebox{\textwidth}{!}{%
    \begin{tabular}{cccl}
        \toprule
        Coarse-grained Category & Criteria & Form-level Category & Explanation \\ \midrule
        \multirow{4}{*}{Metaphor} & \multirow{4}{*}{\begin{tabular}[c]{@{}c@{}}The explicitness of \\rhetorical components\end{tabular}} & Simile & \begin{tabular}[c]{@{}l@{}}Tenor, vehicle and comparator are used \\ explicitly in the sentence.\end{tabular} \\
         &  & Metaphor & \begin{tabular}[c]{@{}l@{}}Tenor and vehicle are used explicitly \\ in the sentence.\end{tabular} \\
         &  & Metonymy & \begin{tabular}[c]{@{}l@{}}Only vehicle is used explicitly in the \\ sentence.\end{tabular} \\ \midrule
        \multirow{7}{*}{Personification} & \multirow{7}{*}{\begin{tabular}[c]{@{}c@{}}The parts of speech of\\rhetorical components\end{tabular}} & Noun & \begin{tabular}[c]{@{}l@{}}Use nouns for people/objects to describe \\ objects/people.\end{tabular} \\
         &  & Verb & \begin{tabular}[c]{@{}l@{}}Use verbs for people/objects to describe \\ objects/people.\end{tabular} \\
         &  & Adjective & \begin{tabular}[c]{@{}l@{}}Use adjectives for people/objects to \\ describe objects/people.\end{tabular} \\
         &  & Adverb & \begin{tabular}[c]{@{}l@{}}Use adverbs for people/objects to describe \\ objects/people.\end{tabular} \\ \midrule
        \multirow{4}{*}{Hyperbole} & \multirow{4}{*}{\begin{tabular}[c]{@{}c@{}}The form of\\hyperbole\end{tabular}} & Direct Hyperbole & Directly exaggerate something. \\
         &  & Indirect Hyperbole & \begin{tabular}[c]{@{}l@{}}Exaggerate something else to exaggerate \\ a thing.\end{tabular} \\
         &  & Mixed Hyperbole & \begin{tabular}[c]{@{}l@{}}Exaggerate using other rhetorical~devices.\end{tabular} \\ \midrule
        \multirow{3}{*}{Parallelism} & \multirow{3}{*}{\begin{tabular}[c]{@{}c@{}}The component of\\parallelism item\\\end{tabular}} & Structure Parallelism & \begin{tabular}[c]{@{}l@{}}The item servers as a specific\\grammatical component in the sentence.\end{tabular} \\
         &  & Sentence Parallelism & \begin{tabular}[c]{@{}l@{}}The item servers as a complete\\sentence on its own.\end{tabular} \\ \bottomrule
    \end{tabular}
    }
    \caption{Annotation guidelines for fine-grained form-level categories in \mydataset.}
    \label{tab:guideline-fc}
\end{table*}

\begin{table*}[!htb]
    \centering
    \resizebox{\textwidth}{!}{%
    \begin{tabular}{cccl}
        \toprule
        Coarse-grained Category & Criteria & Content-level Category & Explanation \\ \midrule
        \multirow{3}{*}{Metaphor} & \multirow{3}{*}{\begin{tabular}[c]{@{}c@{}}The property of\\tenor\end{tabular}} & Concrete & The tenor can be seen, touched or imagined. \\
         &  & Action & The tenor is an action, behavior or event. \\
         &  & Abstract & The tenor is an abstract concept. \\ \midrule
        \multirow{3}{*}{Personification} & \multirow{3}{*}{\begin{tabular}[c]{@{}c@{}}The property of\\content\end{tabular}} & Personification & Write about a non-human as if it were human. \\
         &  & Anthropomorphism & \begin{tabular}[c]{@{}l@{}}Write about something that is not A as if it \\ were A, where A is non-human.\end{tabular} \\ \midrule
        \multirow{3}{*}{Hyperbole} & \multirow{3}{*}{\begin{tabular}[c]{@{}c@{}}The direction of\\hyperbole\end{tabular}} & Amplification & Exaggeration towards large, many, long or high. \\
         &  & Understatement & Exaggeration towards small, few, short or low. \\
         &  & Prolepsis & Mentioning a later event before an earlier event. \\ \midrule
        \multirow{5}{*}{Parallelism} & \multirow{5}{*}{\begin{tabular}[c]{@{}c@{}}The relationship\\between items\end{tabular}} & Coordination & \begin{tabular}[c]{@{}l@{}}Changing the order of the items does not affect \\ the coherence.\end{tabular} \\
         &  & Subordination & \begin{tabular}[c]{@{}l@{}}A logical order of precedence between items \\ exists.\end{tabular} \\
         &  & Gradation & \begin{tabular}[c]{@{}l@{}}The meanings and emotions expressed by each \\ item progressively intensify.\end{tabular} \\ \bottomrule
    \end{tabular}
    }
    \caption{Annotation guidelines for fine-grained content-level categories in \mydataset.}
    \label{tab:guideline-cc}
\end{table*}

Additionally, the annotation guidelines for rhetorical components are shown in Table \ref{tab:guideline-ce}. As mentioned in Section \ref{sec:dataset-view}, we abstract the rhetorical components into three types: connectors, objects and contents. Specifically, for different coarse-grained categories or fine-grained form-level categories, the rhetorical components have various meanings:

\begin{table*}[!htb]
    \centering
    \resizebox{\textwidth}{!}{%
    \begin{tabular}{clcccc}
        \toprule
        \multirow{2}{*}{\begin{tabular}[c]{@{}c@{}}Coarse-grained \\Category\end{tabular}} & \multirow{2}{*}{Criteria} & \multirow{2}{*}{\begin{tabular}[c]{@{}c@{}}Form-level \\Category~\end{tabular}} & \multicolumn{3}{c}{Rhetorical Components} \\ \cmidrule{4-6}
         &  &  & Connector & Object & Content \\ \midrule
        \multirow{3}{*}{Metaphor} & Tenor: the object or concept being compared & Simile & Comparator & \multirow{2}{*}{Tenor} & \multirow{3}{*}{Vehicle} \\
         & Vehicle: the object or concept used for comparison & Metaphor & - &  &  \\
         & Compartor: the word connects the tenor and vehicle & Metonymy & - & - &  \\ \midrule
        \multirow{2}{*}{Personification} & Personification Object: the person/thing being described & \multirow{2}{*}{-} & - & \multirow{2}{*}{\begin{tabular}[c]{@{}c@{}}Personification\\Object\end{tabular}} & \multirow{2}{*}{\begin{tabular}[c]{@{}c@{}}Personification\\Content\end{tabular}} \\
         & Personification Content: the similarities to the object &  & - &  &  \\ \midrule
        \multirow{2}{*}{Hyperbole} & Hyperbole Object: the thing being described & \multirow{2}{*}{-} & - & \multirow{2}{*}{\begin{tabular}[c]{@{}c@{}}Hyperbole\\Object\end{tabular}} & \multirow{2}{*}{\begin{tabular}[c]{@{}c@{}}Hyperbole\\Content\end{tabular}} \\
         & Hyperbole Content: the exaggerated description &  & - &  &  \\ \midrule
        Parallelism & Parallelism Item: the markers & - & \begin{tabular}[c]{@{}c@{}}Parallelism\\Marker\end{tabular} & - & - \\ \bottomrule
    \end{tabular}%
    }
    \caption{Annotation guidelines for rhetorical components in \mydataset.}
    \label{tab:guideline-ce}
\end{table*}

\begin{itemize}
    \item For metaphor, if the form-level category is simile, the rhetorical components include the comparator (as the connector), the tenor (as the object) and the vehicle (as the content). If the form-level category is metaphor, the rhetorical components include the tenor (as the object) and the vehicle (as the content). If the form-level category is metonymy, the rhetorical components only include the vehicle (as the content).
    \item For personification, regardless of the form-level category, the rhetorical components include the personification object (as the object) and the personification content (as the content).
    \item For hyperbole, regardless of the form-level category, the rhetorical components include the hyperbole object (as the object) and the hyperbole content (as the content).
    \item For parallelism, regardless of the form-level category, the rhetorical components only include the parallelism marker (as the connector).
\end{itemize}

\subsection{Details of Dataset Annotation Process}
\label{sec:anno-proc-details}
\begin{figure}[!htb]
    \centering
    \includegraphics[width=\hsize]{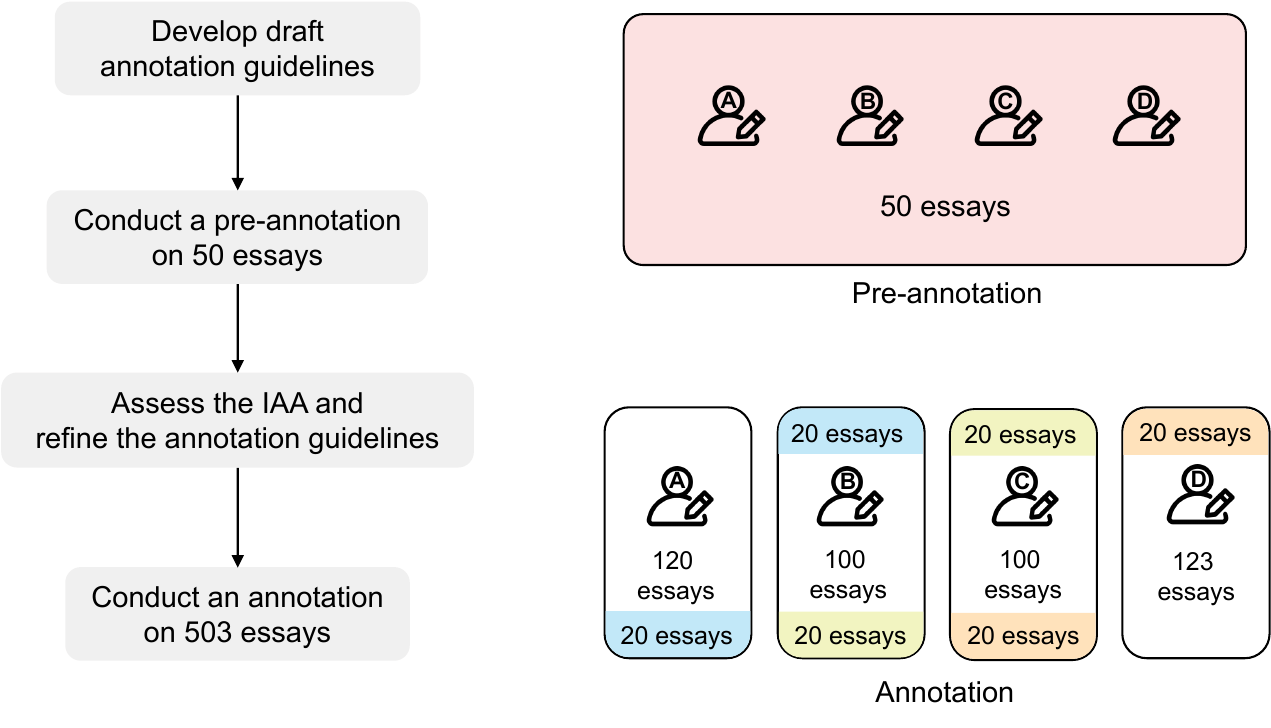}
    \caption{Annotation process of \mydataset.}
    \label{fig:anno-proc}
\end{figure}

The annotation process is illustrated in Figure \ref{fig:anno-proc} and introduced briefly in Section \ref{sec:dataset-proc}. In this section, we further discuss more details of the annotation process.

The entire annotation process, from developing the draft annotation guidelines to conducting an annotation on 503 essays, took three months. To ensure the efficiency and quality of annotation, we held weekly online discussions to address common issues encountered during both the pre-annotation on 50 essays and the annotation on 503 essays. Furthermore, the 50 essays annotated during the pre-annotation process were not re-annotated or used subsequently. We examined the IAA \citep{cohen1960coefficient} of the pre-annotation to identify the challenges we faced during this process. As shown in Table \ref{tab:iaa-pre-anno}, we found that during pre-annotation on sentences employing personification and hyperbole, the IAA was relatively low compared to the other two coarse-grained categories. Therefore, we conducted discussions on these two categories to improve the quality of the annotation.

\begin{table}[!htb]
    \centering
    \begin{tabular}{cc}
      \toprule
      \multicolumn{1}{c}{\multirow{2}{*}{\begin{tabular}[c]{@{}c@{}}Coarse-grained\\ Categories\end{tabular}}} & \multicolumn{1}{c}{\multirow{2}{*}{\begin{tabular}[c]{@{}c@{}}Avg.\\ Cohen's Kappa \(\kappa\) (\%)\end{tabular}}} \\
\multicolumn{1}{c}{} & \multicolumn{1}{c}{} \\ \midrule
      % Coarse-grained Categories & Avg. Cohen’s Kappa \(\kappa\) (\%) \\ \midrule
      Metaphor & 37.07 \\
      Personification & 23.43 \\
      Hyperbole & 23.46 \\
      Parallelism & 27.66 \\ \bottomrule
      \end{tabular}
    \caption{Average Inter-Annotator Agreements across four coarse-grained categories during the pre-annotation.}
    \label{tab:iaa-pre-anno}
\end{table}

\section{Dataset Statistics Details}
\label{sec:data-stat-details}
The statistics of essays used to construct \mydataset are shown in Table \ref{tab:essay-stat}. The total number of sentences in 503 essays is 10,349, with 355,352 tokens.

\begin{table}[!htb]
    \centering
    \begin{tabular}{ll}
      \toprule
       \#Total Sentences  & 10,349 \\
       \#Total Tokens  & 355,352 \\
       Avg. \#Sentences per Essay & 20.57 \\
       Avg. \#Tokens per Essay & 706.47 \\
       Avg. \#Tokens per Sentence & 34.34 \\ \bottomrule
    \end{tabular}
    \caption{Statistics of essays used to construct \mydataset.}
    \label{tab:essay-stat}
\end{table}

\section{Experimental Setups}
\label{sec:exp-setups}

We split \mydataset into training/validation/test sets, displayed in Table \ref{tab:data-split}. To prevent data leakage, the dataset is split at the essay level, ensuring that the essays containing sentences in the training or validation sets are not included in the test set for any task.

\begin{table}[!htb]
    \centering
    \resizebox{0.88\hsize}{!}{%
    \begin{tabular}{@{}cccc@{}}
    \toprule
    Tasks & Type & \#Sentences & \#Tokens \\ \midrule
    \multirow{4}{*}{RC/FC/CC/CE} & Train & 634 & 29,517 \\
     & Val & 225 & 11,748 \\
     & Test & 248 & 12,186 \\
     & \textbf{Sum} & \textbf{1,107} & \textbf{53,451} \\ \midrule
    \multirow{4}{*}{RG} & Train & 404 & 52,969 \\
     & Val & 158 & 22,246 \\
     & Test & 169 & 24,239 \\
     & \textbf{Sum} & \textbf{731} & \textbf{99,454} \\ \bottomrule
    \end{tabular}%
    }
    \caption{Dataset splits of \mydataset.}
    \label{tab:data-split}
\end{table}

We perform full parameter fine-tuning of RoBERTa on 24GB RTX 3090 GPUs and LoRA \citep{hu2021lora} fine-tuning of Qwen1.5-7B on 80GB A100 GPUs. The results from automatic evaluation metrics are averaged over three runs with different random seeds. The hyperparameters used in our experiments are listed in Table \ref{tab:hyperparams}. Our models are fine-tuned using AdamW \citep{loshchilov2017decoupled} optimizer and cosine learning rate scheduler.

\begin{table}[!htb]
    \centering
    \resizebox{\hsize}{!}{%
    \begin{tabular}{lccccc}
      \toprule
      Models & lr & bs & steps & r & \(\alpha\)  \\ \midrule
        RoBERTa & \(6\times10^{-5}\) & 32 & 30 epochs & - & - \\
        Qwen1.5-7B\textsubscript{Single} & \(2\times 10^{-4}\) & 32 & 50 steps & 32 & 32 \\
        Qwen1.5-7B\textsubscript{Multi} & \(2\times 10^{-4}\) & 32 & 250 steps & 32 & 32\\ \bottomrule
    \end{tabular}%
    }
    \caption{Hyperparameters for fine-tuning RoBERTa and Qwen1.5-7B. "lr" refers to the learning rate. "bs" refers to the batch size. "r" and "\(\alpha\)" refer to the hyperparameters used in LoRA.}
    \label{tab:hyperparams}
\end{table}

\section{Prompt Templates}
\label{sec:prompt}
\begin{figure*}[!htb]
    \centering
    \includegraphics[width=\textwidth]{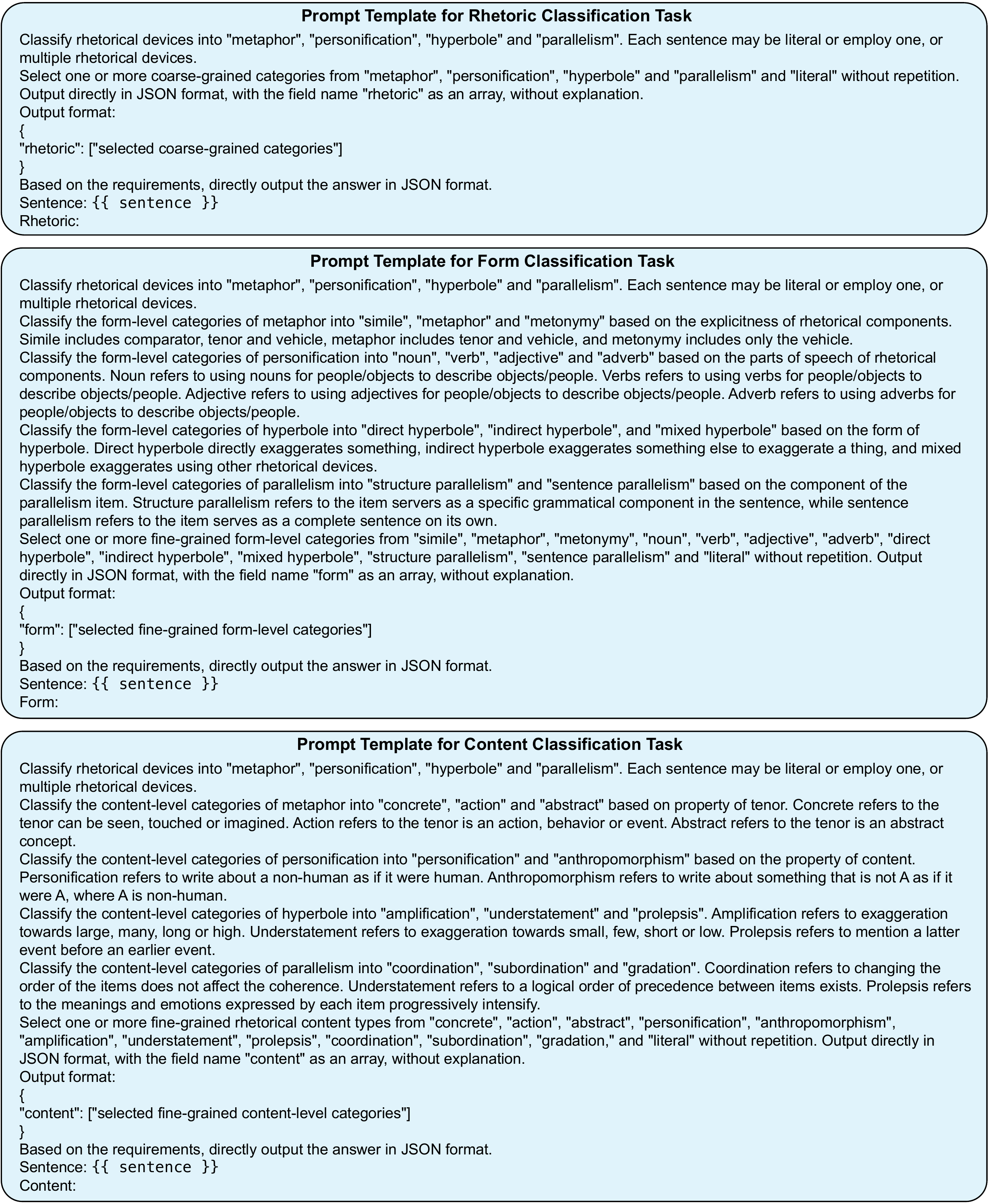}
    \caption{Prompt templates for Tasks RC, FC and CC. \texttt{\{\{sentence\}\}} represents the input sentence.}
    \label{fig:prompt-classification}
\end{figure*}

\begin{figure*}[!htb]
    \centering
    \includegraphics[width=\textwidth]{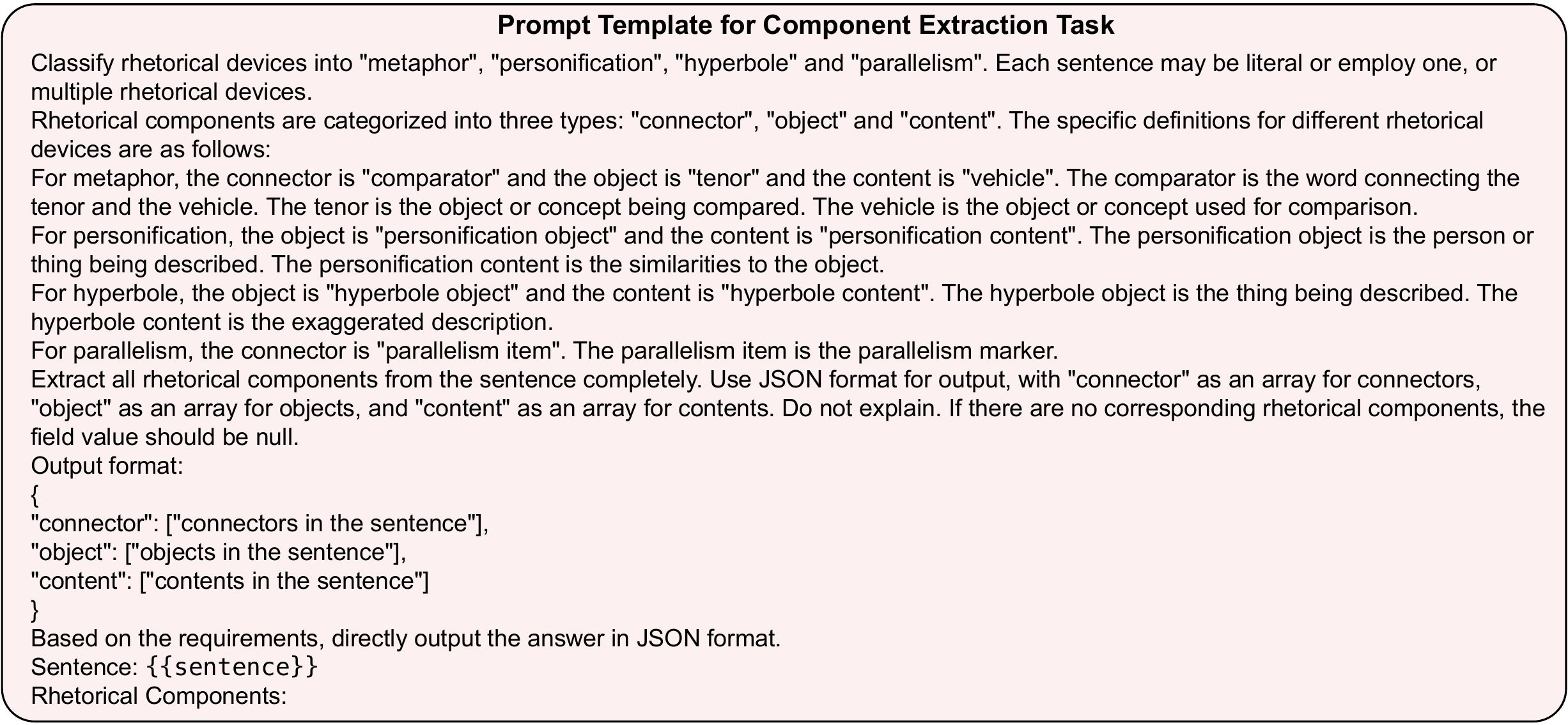}
    \caption{Prompt template for Tasks CE. \texttt{\{\{sentence\}\}} represents the input sentence.}
    \label{fig:prompt-extraction}
\end{figure*}

\begin{figure*}[!htb]
    \centering
    \includegraphics[width=\textwidth]{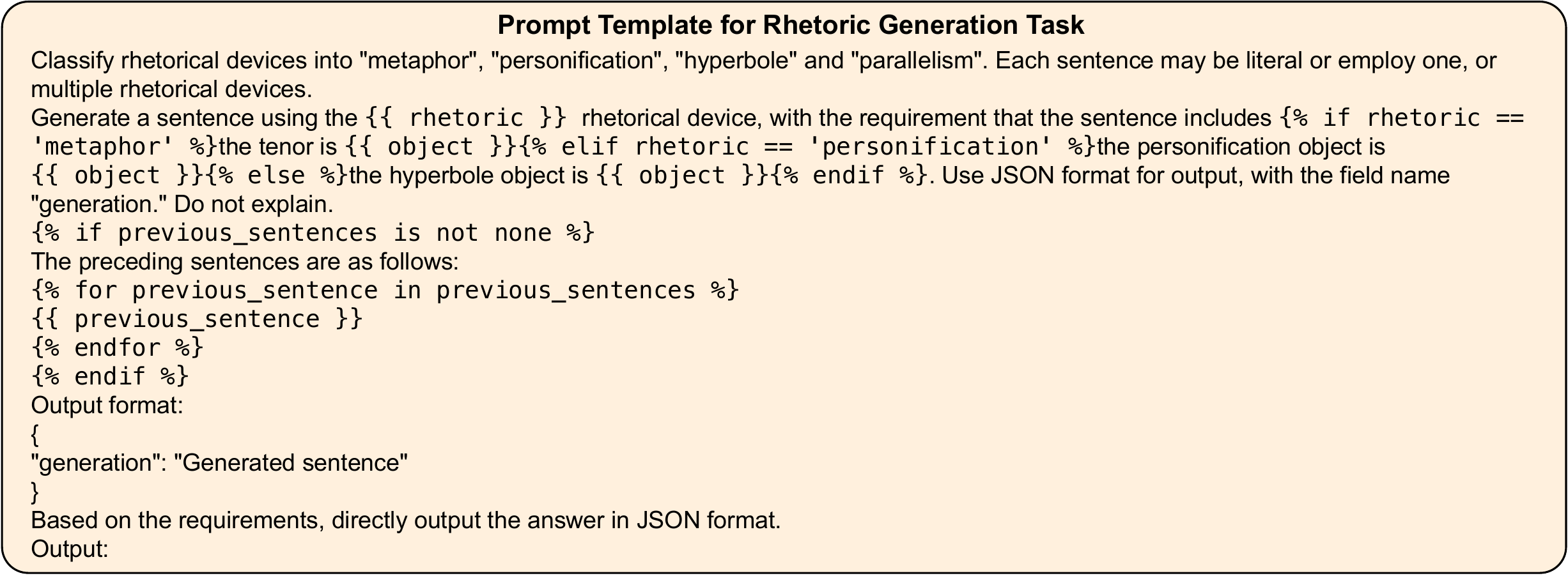}
    \caption{Prompt template for Tasks RG. \texttt{\{\{rhetoric\}\}} represents the target coarse-grained category. \texttt{\{\{object\}\}} represents the target object. \texttt{\{\{previous\_sentence\}\}} represents the preceding context.}
    \label{fig:prompt-generation}
\end{figure*}

For all tasks and models, the prompt templates are used for both inference and fine-tuning. The prompt templates and inputs are originally written in Chinese. The English translations of the prompt templates are displayed in Figure \ref{fig:prompt-classification}, Figure \ref{fig:prompt-extraction} and Figure \ref{fig:prompt-generation} respectively.

\end{document}